# An Overview of Two Age Synthesis and Estimation Techniques

**Milad Taleby Ahvanooey[1]\*, Qianmu Li[1]**

1. School of Computer Science and Engineering, Nanjing University of Science and Technology, Nanjing, P.O. Box 210094 P.R, China. e-mail: taleby@njust.edu.cn, Qianmu.@njust.edu.cn

**Abstract**

Age estimation is a technique for predicting human ages from digital facial images, which analyzes a person's face image and estimates his/her age based on the year measure. Nowadays, intelligent age estimation and age synthesis have become particularly prevalent research topics in computer vision and face verification systems. Age synthesis is defined to render a facial image aesthetically with rejuvenating and natural aging effects on the person's face. Age estimation is defined to label a facial image automatically with the age group (year range) or the exact age (year) of the person's face. In this case study, we overview the existing models, popular techniques, system performances, and technical challenges related to the facial image-based age synthesis and estimation topics. The main goal of this review is to provide an easy understanding and promising future directions with systematic discussions.

## 1. Introduction

Age estimation and face verification have many applications in modern machine vision systems. Technically, age-separated facial images contrast significantly in both texture and shape. In general, the human face has important amount of attributes and features such as gender, expression and age. The most of people can easily recognize human characteristics like emotional filings, where they can guess if the human is sad, angry or happy from the face. In the same trend, it is easy to detect the gender of the human. However, predicting human age just by looking at recent or old pictures, is an unpredictable challenge for modern computer vision systems in new daily life. Commonly, the facial aging attributes depend on many impressive factors such as degree of stress and life style. For example, smoking causes several facial attributes changes [1-3].

During the last two decades, many researches and survey papers have been written to introduced various aspects of age estimation and facial aging techniques. Some provide profound introductions to the age estimation and modeling as a whole. However, there has been written no new overview on computer-based age synthesis and estimation in recent years, and most of the newcomers in computer vision aim to learn about age estimation techniques due to their explosively emerging real-world applications, such as forensic art, electronic customer relationship management, entertainment, security control and surveillance monitoring, biometrics, and cosmetology, which is caused to draw towards writing an easy overview to help their understanding about the facial age estimation[1-6]. This overview study aims to fill the gap, by providing an easy guide to understanding & interpreting the age estimation techniques for both newcomers and researchers.

The rest of this overview is organized as follows. Section (2) presents a brief description of age synthesis on faces, age estimation system with systematic discussions. Section (3) summarizes the performance of some highlight techniques and suggests some guidelines and research directions. Finally, section (4) draws some conclusions.

## 2. Literature Review

Technically, the human age estimation is a non- reversible process due to existing different humane face characteristics that change during the human lifetime such as hair whitening, muscles drop, and wrinkles, etc., even though using beauty care products may slightly invert minor photo-aging effects. Generally, people have various models of aging, with time human faces begin to get different styles in different ages, and there are general characteristics which can be extracted during the estimation process [3]-[7]. The existing literature can be classified into two main stages: Age synthesis (or age progression) and age estimation.

### 1-2- Age Synthesis on Faces

Age synthesis or age progression is the generic process of building a face model. Face modeling has been prevalent for a long time in both "machine vision" and "computer graphics" [8]-[17]. For example, Fig.1 shows a facial modeling which focuses on facial zones and analysis of wrinkles for each zone.

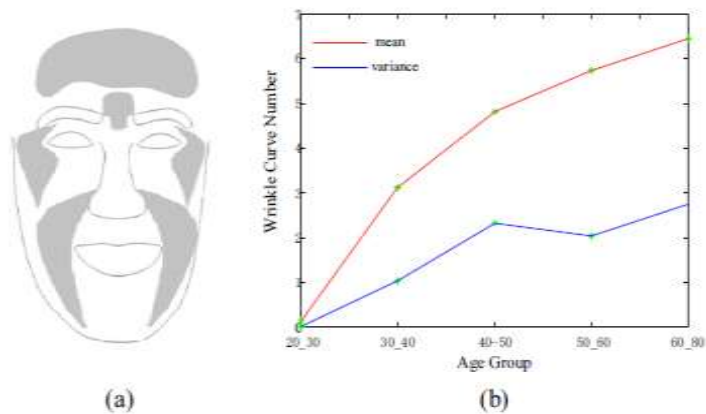

**Figure 1. An example of face modeling "Facial zones and statistic of forehead wrinkle number" [14]**

The face model is depicted in figure above, was proposed by Sou et al. in [14]. In the model, they have decomposed a face into 8 zones, as highlighted in figure.1 (a), and added the wrinkles for each zone separately. Due to collecting the wrinkle images of each person across a sequence of age groups is impractical, they collected a large number of face images at different age groups, and utilized the statistical data instead. Moreover, they provided a facial modeling by analyzing the number of wrinkles with Poisson distribution.

Cootes et al. [8] have presented a technique for producing statistical models from a set of training examples. During the process, they performed Principal Component Analysis (PCA) on the deviations of each example from the mean example. As a result of their analysis, the training examples can be reconstructed by equation 1.

$$X = X_m + P_b \qquad (1)$$

Where $X$ is a vector expressing the intensity pattern or shape of a training example (face image), $X_m$ is the mean example, $P$ is the matrix of eigenvectors, and $b$ is a vector of model parameters or weights [18].

Edwards et al. [9] have explained in detail how the statistical model can be applied for modeling merged shape and intensity variation in the facial images. Later on, Edwards et al. [10] have advanced their previous model [9] in order to describe how color models can be produced by consisting of the intensity pattern (e.g., the RGB component of each pixel in the facial regions).

Recentlly, age progression is the one of the most commonly used tool for forensic purposes by law enforcement. This technique has been applied for enhancing a photograph in order to identification of victim, lost person, etc. so far. Moreover, this technique can be employed to design the likely current appearance of a missing person from his facial photos (many years old). Age synthesis by machine can significantly increase the efficiency of practical works while providing much photo-realistic aging effects that can achieve the needs of aesthetics.

In the following, we have discussed two popular synthesis frameworks consisting: Implicit statistical synthesis and explicit mechanical synthesis in the following points.

**1-1-2- Implicit Statistical Synthesis**

The implicit synthesis usually utilizes statistical methods and focuses on the appearance analysis by considering shapes and texture synthesis simultaneously. This synthesis requires to collect a database which consists a large number of facial images with a various range of ages. In case of the aging space, each face image is considered as a high-dimensional point and, then, the age synthesis can be animated by drawing the distances between faces with various ages or the model parameters manipulating different appearance changes [1-5]. In this section, we summarized some existing techniques that are based on implicit statistical synthesis.

Ramanathan and Chellappa [13] have developed a twofold approach for modeling facial aging in adult. In this work, a shape transformation model is expressed as a physically-based parametric muscle model, while the texture aging is operated by an image wrinkle extraction function based on gradient analysis. This model is able to simulate and estimate facial aging in two ways: weight loss and weight-gain. The wrinkle simulation module can produce various effects such as moderate, strong, and subtle.

Lanitis et al. [11] have employed the active appearance model (AAM) method [12], to generate the aging functions from a specific data set (e.g., young faces under 30 years old), in which a PCA is used to extract the texture and shape variations. The PCA coefficients for the linear alteration of training elements considered as model parameters and constrain various types of appearance variations. This model can also be employed during the optimization and age normalization for improving face recognition performance. Figure.2 demonstrates the aging appearance simulation result and an example of AAM by Lanitis method.

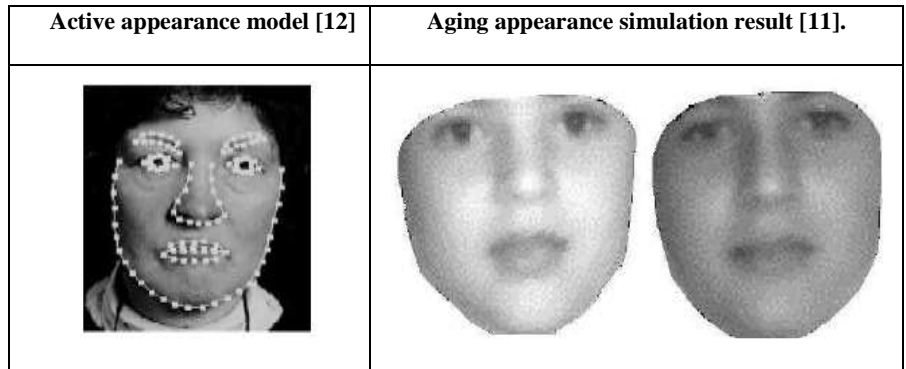

**Figure2. Face aging using the PCA and the AAM**

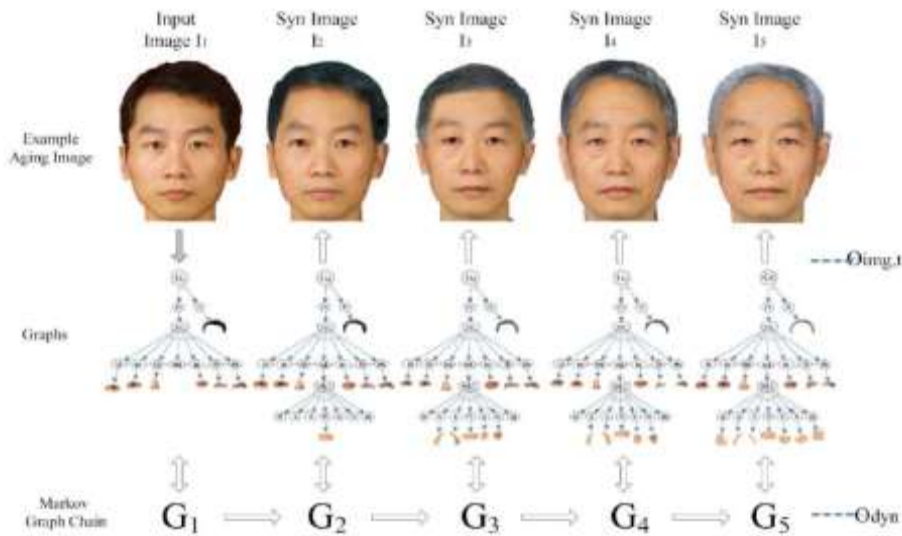

**Figure.3. Multi-layer dynamic face aging model and results [14]**

Suo et al. [14] have introduced a dynamic face aging framework based on multi-resolution and multi-layer image representations. In this model, the framework represented a set of 50.000 facial images by a multi-layer And-Or graph and integrated three most prominent aspects akin to aging changes: global appearance alterations in shape and hair style, deformations and aging effects of facial elements, and wrinkles appearance at different facial zones. A process model based on dynamic Markov is generated on the graph structures of the training dataset. Therefore, these structures over various age categories are finally tested in terms of the dynamic model to simulate new aging faces. Figure.3 depicts the dynamic face aging model with some simulation results that demonstrate the photorealistic results.

Fu and Zheng [15] have been presented a novel facial modeling framework (M-Face) based on appearance photorealistic of the multiple facial object class for emotional face rendering. The Aging Ratio Image (ARI) is obtained from their database in order to texture rendering. In this work, M-Face integrates texture MRI-mapping and the 2-D shape caricaturing techniques for bypassing the 3-D face reconstruction. According to the driven by simplified model parameters and 2-D face examples, the representation morphing, chronological aging and illumination variance can be combined seamlessly in a photorealistic shape on the view-rotated faces. Practically, M-Face system has been successfully evaluated for criminal appearance sketching by the police department of Henan province in China. Figure.4 illustrates an example of M-Face process.

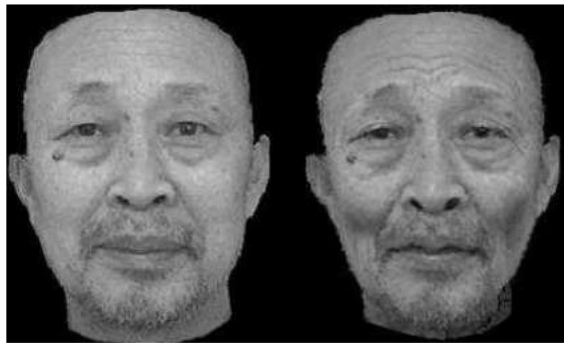

**Figure.4. An example of M-Face rendering [15]**

**2-1-2- Explicit Mechanical Synthesis**

Basically, explicit mechanical synthesis functions based on the texture analysis, which is further akin to skin aging, the most different facial changes after adulthood. Moreover, during the skin aging, wrinkles appear and become more stereometric because of the nature of skin and muscle compression. This synthesis is advanced using image-based rendering for the aim of photorealistic appearance prevision across age progression. Let's suppose that, given a face image for rendering, this technique either builds a model for capturing aging patterns or aging synthesis from sample faces and clones to a young face [1]- [5].

Shan et al. [16], [17] have proposed an image-based surface detail transfer (IBSDT) technique for transferring geometric structures through two object surfaces. The researchers achieved that geometric details can be obtained without knowing the surface reflectance. Moreover, the geometric structures can be propagated for rendering to another surfaces. In practice, this technique can employ to face aging, simulating old age, and removing wrinkles of facial images.

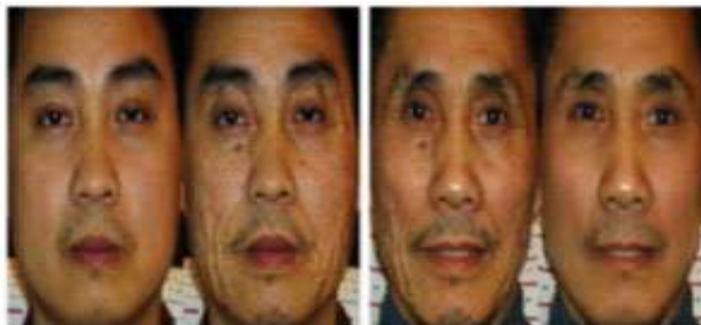

**Figure5. Results of face aging and simulating old face of the original photos using IBSDT [16], [17]**

As depicted in Figure.5, the first and third columns demonstrate the original images of two persons with different age. The second and fourth columns demonstrate the two rendering results for face aging and simulating old age by the IBSDT technique (i.e., exchange the age features between the two faces).

**2-2- Age Estimation System**

Computationally, an age estimation system includes of two modules: image representation and age estimation.

**1-2-2- Image Representation**

   a) **Anthropometric models**

Kwon and Lobo [18], [19] have presented a theory and practical computations for visual age classification from facial images. This theory utilizes a mathematical model to express the growth of a person's head from infancy to adulthood. Farkas [20] has introduced an exhaustive overview of face anthropometry which is the science of proportions and measuring sizes on the human faces. In general, people employs the distance ratios measured from facial landmarks for identifying age growth characterization, instead of applying directly the mathematical models. In fact, there are two reasons that people refuse to apply the mathematical formulation for age estimation: (I) it is hard to measure the pattern of face from 2D facial images and the mathematical model cannot analyze the growing face naturally, while the ages are close to adults [13]. In addition, they have categorized human faces into three classifications: babies, young adults, and senior adults. The authors examined their technique on a small dataset consisting of 47 faces, but they did not report the overall efficiency of the technique on this dataset. In practice, age estimation techniques based on the anthropometry model can only deal with young ages, due to during the adult period, human face style does not change too much considerable. They also computed wrinkles between different face images to separate young adults from senior adults. The wrinkles were analyzed based on various regions of faces such as on the forehead, next to the eyes, and near the cheekbones, etc. Based on our finding, there are no reported results on a large dataset for age estimation using the anthropometry model. Therefore, since the anthropometric model only consider the facial geometry without any texture information, and the anthropometric measurements, it might be useful for young ages, but it is not proper for adults.

   b) **Active appearance models**

This model is a statistical-based face model was presented initially in [12] for coding face images. In that study, given a set of training facial images, a statistical shape model and an intensity model are learned separately, according to the principal component analysis which he AAMs were employed successfully for face encoding. Lanitis et al. [11] have advanced the AAMs for face aging by presenting an aging function, $age=f(b)$ to describe the variation in age. In the function, age is the real age of a person in a face image, b is a vector including 50 raw model parameters learned from the AAMs, and f is the aging function. In fact, this function explains the relationship between the age of persons and the parametric structure of the face images. The experiments were executed on a dataset consisting of 500 face images of 60 subjects.

Lanitis et al. [21] have also attempted various classifiers for age estimation using their age image representation based on the quadratic aging function or WAS (Weighted Appearance Specific) [11]. In practice, the AAMs can deal with different ages in general, rather than just young ages compared to the anthropometry model based techniques [13], [18], [19]. Moreover, the AAMs based techniques analyze both the shape and texture rather than just the facial geometry as in the anthropometric model based methods during the age image representation. These techniques can apply to the precise age estimation, since each test image will be labelled with a unique age value selected from a continuous range. The further optimizations on these aging pattern representation techniques are: (I) to present evidence that the relationship between face and age can be essentially expressed by a quadratic function; (II) to consider outliers in the age labeling space; and (III) to handle high-dimensional parameters.

   c) **Aging Pattern Subspace**

In this model, a sequence of a person's aging facial images used all together as model during the aging process. This model was suggested by Geng et al. [22], [23], which is named AGing pattErn Subspace (AGES). In this technique, they defined an aging pattern as a sequence of personal face images such that all images are belonged to the same person, sorted in a temporal order. If the facial images of all face ages be available for a person, then the related aging pattern will call a perfect aging pattern; else, it will call an unperfect aging pattern. The AGES technique is able to simulate the missing ages by applying an EM-like iterative learning algorithm, which works based on two phases: the learning phase and the age estimation phase. In learning of the aging pattern subspace, the PCA was employed to achieve a subspace representation. In comparison to the standard PCA, there are possibly some missing age images for each aging pattern. The EM-like iterative learning technique is applied to minimize the reconstruction error identified by the difference among the reconstructed face images and the available age images. The basic values for missing faces are set by the average of available face images. Thus, the eigenvectors of the covariance matrix of all facial images and the means of faces can be calculated. The process iterates until the reconstruction error is to reach the minimum enough size. In the age estimation stage, the test face image requires to discover an aging pattern fitting for it and an appropriate age position in that aging pattern. Therefore, the age position is fetched as the estimation of the age of the test face image. To meet that requirement, the test face is confirmed at every feasible position in the aging pattern and the one with the minimum reconstruction error is chosen.

In case of applying the AAMs for encoding facial images, Geng et al. have employed 200 AAMs characteristics to encode each face image [22], [23]. They also evaluated the AGES technique by performing on the FG-NET aging database [24] that reported the Mean Absolute Error (MAE) as 6.77 years. Practically, there is a problem in use of the AGES technique, when it estimates the age of an input face image, the AGES technique supposes, there exist facial images of the same person but at various ages or at least a similar aging pattern for the specific face image in the training database. This assumption might not be satisfied for some aging databases. And also it is really hard to collect a large database including facial images of the same person at many various ages with close imaging characteristics. Another problem of the AGES technique is that the AAMs face representation might not encode facial wrinkles good for senior people, due to the AAMs technique only encodes the image intensities without applying any spatial neighborhood to compute texture patterns (e.g., the intensity of an individual pixel may not characterize it adequately). To represent the facial wrinkles for older persons, the texture patterns at local regions are required to be considered.

### d) Appearance Models

Age estimation based on appearance feature is more focused by the appearance model(AM). In this model, two type of feature extraction were applied in existing age estimation techniques.

Hill et al. [26] have utilized the Local Binary Patterns (LBP) [25], for appearance feature extraction in an automatic age estimation technique. Therefore, this technique obtained 80% accuracy on the FERET database with nearest neighbor classification, and 80-90% on the FERET and PIE databases [28].

Yan et al. [29], [30] have suggested a feature descriptor using Spatially Flexible Patch (SFP). In addition, since the SFP considers local patches and their position information, face images with small occlusion, misalignment, and head pose alterations can still be manipulated effectively. In addition, it can also enhance the discriminating property of the feature set when deficient samples are given. In practice, it was Modeled by a Gaussian Mixture Model (GMM), which achieved a MAE of 4.95 years on the FG-NET aging database [24], and MAEs of 4.38 years and 4.94 years for male and female on the YGA database respectively.

Suo et al. [31], have introduced an age estimation technique based on hierarchical face model for faces appearing at low, middle and high resolution respectively. Basically, within the hierarchical model, they used specific filters to different parameters at different levels for feature extraction. This technique achieved a MAE of 5.974 years on the FG-NET aging database using Multi-Layer Perceptron (MLP), and MAE of 4.68 years (with error tolerance of 10 years' estimation rate attains 91.6%) on their own database respectively.

Guo et al. [32] have provided a new technique using the Biologically Inspired Features (BIF) [33], [34] for age estimation via faces. BIF obtained MAE of 4.77 years on the FG-NET aging database, and MAEs of 3.91 years and 3.47 years for female and male on the YGA database respectively. In [35], Guo et al. have studied age estimation problems by considering both age and gender estimation in a technique. In this work, the BIF+Age Manifold feature merged with the support vector machine (SVM) which achieved MAEs of 2.61 years and 2.58 years for female and male on the YGA database respectively. These results confirm the superiority of the BIF for the age estimation area.

Luu et al. [36] have proposed a novel Contourlet Appearance Model (CAM) which is more precise and faster at localizing facial landmarks than the AAMs. The CAM obtained MAEs of 4.12 years on the FG-NET aging database, and MAEs of 6.0 on the PAL data base.

**2-2-2- Age Estimation**

After an aging feature representation (or extraction), the next step is to estimate ages. Basically, age estimation techniques have two main categories: a) regression-based; a) classification-based.

a) **Classification-based**

Anitis et al. [21] have analyzed the performance of different classifiers for age estimation techniques, consisting the nearest neighbor classifier, the Artificial Neural Networks (ANNs), and a quadratic function classifier. This framework utilizes the AAMs in order to represent facial images. In the experimental results, they have tested their framework on a small database including 400 images at ages ranging from 0 to 35 years and, moreover, they claimed that the quadratic function classifier can achieve 5.04 years of MAE, that is lower than the nearest neighbor classifier, but higher than the ANNs.

In [44], formulated as an 11-class classification problem for the WIT-DB database (separate male and female), Ueki et al. have constructed "11" Gaussian models in a low-dimensional 2DLDA+LDA feature space using the EM technique. The classification age-group is characterized by fitting the test image to each Gaussian model and comparing the likelihoods. This technique achieved accuracies of about 50% for male and 43% for female on the 5-year range age-group classification and obtained accuracies of about 72% for male and 63% for female on the 10-year range age-group classification. For 15-year range age-group classification, it obtained accuracies of about 82% for male and 74% for female.

Wang et al. [27] have presented a novel facia aging technique, which is called "raSVM+". This method discriminates the relative age group during and after find a proper facial feature to express the age attributes for age estimation by analyzing and comparing different feature extraction methods. Moreover, they achieved the rank relationships of the estimation process by registering privileged information in the faces and applying the relative characteristics learning algorithm. The privileged characteristics supplied inherent relationships between various age groups. This technique employed relative attributes as privileged data for raSVM+ learning, which improved the accuracy of age estimation by controlling outliers in the training datasets. The raSVM+ obtained MAEs of 4.07 years on the FG-NET database.

b) **Regression-based**

Lanitis et al. [11] have introduced three formulations for providing the aging function: linear, quadratic, and cubic, respectively, by 50 raw model parameters. Theses parameters are learned from training facial images of various ages based on a genetic algorithm.

In [37], Yan et al. have proposed a semidefinite programming (SDP) for automatic designing regressor, which is an effective tool but its computational is very high. When the size of the training set be large, the solution to SDP may be hard to obtain.

In [38], a convergence provable Expectation-Maximization (EM) method has been applied to solve the regression problem and speed up the optimization process. This method achieved MAEs of 6.95 years for both male and female on the YGA database, and 5.33 for the FG-NET aging database.

Suo et al. [31] have compared the Age group specific Linear Regression (ALR), Support Vector Regression (SVR), MLP and logistic regression (multi-class Adaboost [40]) on FG-NET and their own databases. Therefore, they obtained the best efficiency with MLP in the experimental results.

Guo et al. [41], have suggested a locally adjusted robust regressor (LARR) for learning and prediction of human ages. In that study, the researchers utilized the SVMs as a representative classifier, and the SVR as a representative regressor. In addition, they have tested both of classifier and regressor on the same database and compared their efficiency together. In the experimental results, the SVR execute much worse than the SVMS on the YGA database (e.g., 7.0 versus 5.55, and 7.47 versus 5.52 years, for females and males, respectively), while the SVR execute much better than the SVMs on the FG-NET database (5.16 versus 7.16 years). These results confirm that the classification-based age estimation can achieve much better (or worse) than the regression-based techniques in various cases. Later on [42], Guo et al. have attempted to obtain better performance on the LARR by combining a classifier and a regressor. Using the combination technique, the MAEs reached 5.07 years on the FG-NET database, and 5.25 and 5.30 years for female and male on the YGA aging database, respectively.

Zheng et al. [43] have presented a novel facial age estimation technique based on partial least squares regression with label distribution learning (PLS-LLD). The technique transforms the label distribution learning task of facial age estimation into the multivariate multiple regression analysis. In the experimental results, the PLS-LLD achieved MAEs of 3.29-4.55 years on the FG-NET database.

**c) Hybrid Techniques**

As we have already pointed out in above, the age estimation problem can be figured out as either a regression or classification problem. Which technique is better for age estimation (or facial aging): regression or classification?

To achieve an optimal answer, a researcher might pick out some representative regressor and classifiers for comparing their results on the same databases. Guo et al. [41], [42] have employed the SVM as a representative classifier and the SVR as a representative regressor. They compared their performance by performing on the same dataset. In the results, the SVM performs much better than the SVR on the YGA, while the SVM performs much worse than the SVR on the FG-NET database. From their results, we can realize that, the best way for robustness is to combine the regression and classification techniques to take advantage of the merits from both of those techniques.

Liang et al. [45] have introduced two novel classifiers, sequence k-nearest neighbor (SKNN) and ranking-KNN, for age estimation by combining age grouping and age value calculation. In that study, they applied a sequence KNN algorithms to classify an unknown sample into a unique age group by fusing geometric ration and wrinkle density features. Moreover, appearance features combining geometric ration and ULBP textures are applied to predict the age value under the ranking-KNN framework. In the results, they tested the algorithm on FG-NET database and achieved MAEs of 4.97 years.

Ng et al. [46] have proposed a Multi-Layer Age Regression (MAR) by estimating the face age based on a coarse-to-fine prediction using global and local features. In the first layer, the SVR performs a between-group prediction using the parameters of facial appearance model. In the second layer, a within group estimation is executed using facial appearance model, BIF, Kernel-based Local Binary Patterns (KLBP)and Multi-Scale Wrinkle Patterns (MWP). In the result, they claimed that this technique outperforms the state of the art with a MAE as low as 3.00 years on FERET, and 2.71 years on PAL. In addition, it performs the best on FGNET with MAE of 5.39 years and BIF performs the best on MORPH with MAE of 3.20.

Jadid and Rezaei [47] have presented a novel facial age estimation technique utilizing the combination of Haar wavelet transform and color moment (HWCM) methods to extract full informative and influencing features of the face images. Moreover, they utilized a training step in order to improve features by applying an SVR model, based on the extracted features vector. This technique achieved MAEs of 1.97 years for the FG-NET aging database, and MAEs of 2.43 years for the MORPH.

### d) Juvenile Techniques

The juvenile age estimation is a new facial aging area, have been drawing attention from the computer vision researchers recently. In this area, the researchers focus on anthropometric techniques concerning juvenile aging from the facial images [48], [49]. These studies have performed with the purpose of being able to characterize an age, or age range from the faces of persons present in face images of alleged child pornography (e.g., between 10-19 years old). Due to having different category and specific databases, we cannot compare to facial age estimation techniques. More information can be found in ref [50].

## 3. Summary and Suggestions for the future works

An age estimation technique can be approached by either a regressor or classifier since various databases and systems may be too unbalanced or biased for evaluation the age estimation. But it is not exclusively a regression or classification problem. On the other hand, some different feature extraction problems can be applied to solve the task of facial aging. In order to illustrate the performance of existing techniques and have a safe comparison, we selected those highlight techniques which have been performed on the same popular databases such as the FG-NET, the YAMAHA(YGA), and the MORPH. Table.1. summarizes the different age estimation results obtained by computer vision researchers recently, consisting of the details on each database in use in terms of their MAEs. The decreasing MAEs rate leads to greater the performance of the algorithm. The equation 2 is used as common performance criteria to compare the various age estimation techniques in the existing literature.

$$MAEs = \frac{\sum_{i=1}^{N_t} |\hat{x}_i - x_i|}{N_t} \qquad (2)$$

Where $\hat{x}_i$ is the estimated age for the $i^{th}$ testing samples and $x_i$ is the corresponding ground truth age (the true age), and $N_t$ is the total number of testing samples [18]. The CS is defined as $CS(j) = \frac{N_{e \leq j}}{N} \times 100$, where $N_{e \leq j}$ is the total number of test images on which the technique generates an absolute error no higher than j years.

In the following, we suggest some directions aimed at guiding researchers on the best options to improve the age estimation techniques depending on the appearance feature analysis of facial images. However, we have to note that the directions are general and empirically derived rules of thumb; these suggestions must not be considered rigidly or dogmatically.

**Table.1: Summary and Comparison of Highlight Techniques on Different Databases**

| Technique | Database [51] | Data Description | | | Performance | | | |
|---|---|---|---|---|---|---|---|---|
| | | Subject # | Image# | Label (Years) | Algorithm | MEA (Years) | $CS \leq 10$ (Years) | Accuracy |
| AGES, Geng et al. 2006 [22] | FG-NET | 82 | 1002 | 0-69 | R | 6.77 | ≈ 81% | N/A |
| AGES, Geng et al. 2006 [22] | FG-NET | 82 | 1002 | 0-69 | C | 1.27 | N/A | 40.92% (hit rate) |
| CAM, LUU et al. 2011 [36] | FG-NET | 82 | 1002 | 0-69 | AAM | 4.12 | ≈ 90% | 90% |
| MLP, Sou et al. 2008 [31] | FG-NET | 82 | 1002 | 0-69 | R | 5.97 | N/A | N/A |
| LARR, Gou et al. 2008 [41] | FG-NET | 82 | 1002 | 0-69 | R | 5.07 | ≈ 88% | N/A |
| LARR, Gou et al. 2008 [42] | FG-NET | 82 | 1002 | 0-69 | R & C Modified | 4.97 | ≈ 88% | N/A |
| BIF, Gou et al. 2009 [32] | FG-NET | 82 | 1002 | 0-69 | R (SVR) | 4.77 | ≈ 89% | N/A |
| SFP, Yan et al. 2008 [29] | FG-NET | 82 | 1002 | 0-69 | R(GMM) | 4.95 | N/A | N/A |
| SVM+, Wang et al. 2016 [27] | FG-NET | 82 | 1002 | 0-69 | C & SVM | 4.07 | N/A | N/A |
| HWCM, Jadid and Rezaei 2017 [47] | FG-NET | 82 | 1002 | 0-69 | HWCM (SVR) | 1.97 | N/A | N/A |
| Liang et al. 2014 [45] | FG-NET | 82 | 1002 | 0-69 | SKNN | 4.97 | N/A | N/A |
| MAR, Ng et al. 2017 [46] | FG-NET | 82 | 1002 | 0-69 | SVR | 5.39 | N/A | N/A |
| PLS-LDD, Zheng et al. 2017 [43] | FG-NET | 82 | 1002 | 0-69 | R | 3.29-4.55 | N/A | N/A |
| LARR, Gou et al. 2008 [41] | YGA | 4000(M) 4000(F) | 8000 | 0-93 | R | 5.30 (M) 5.25 (F) | ≈ 83% ≈ 81% | N/A |
| LARR, Gou et al. 2008 [42] | YGA | 4000(M) 4000(F) | 8000 | 0-93 | H | 5.12 (M) 5.11 (F) | ≈ 83% ≈ 82% | N/A |
| BIF, Gou et al. 2009 [32] | YGA | 4000(M) 4000(F) | 8000 | 0-93 | C (SVM) | 3.47 (M) 3.91 (F) | ≈ 88% ≈ 85% | N/A |
| BIF, Gou et al. 2009 [35] | YGA | 4000(M) 4000(F) | 8000 | 0-93 | C (SVM) | 2.58 (M) 2.61 (F) | N/A | 89.7% |
| APM, Yan et al. 2008 [29] | YGA | 4000(M) 4000(F) | 8000 | 0-93 | R (Kernel) | 4.38 (M) 4.94 (F) | N/A | N/A |
| SFP, Yan et al. 2008 [30] | YGA | 4000(M) 4000(F) | 8000 | 0-93 | R(GMM) | 7.82 (M) 8.53 (F) | ≈ 75% ≈ 70% | N/A |
| AGES, Geng et al. 2007 [23] | MORPH | 515 | 1430 (M) 294 (F) | 15-68 | R | 8.83 | ≈ 70% | N/A |
| HWCM, Jadid and Rezaei 2017 [47] | MORPH | 515 | 1430 (M) 294 (F) | 15-68 | HWCM (SVR) | 2.43 | N/A | N/A |
| MAR, Ng et al. 2017 [46] | MORPH | 515 | 1430 (M) 294 (F) | 15-68 | SVR | 3.98 | N/A | N/A |
| SVM+, Wang et al. 2016 [27] | MORPH | 515 | 1430 (M) 294 (F) | 15-68 | C & SVM | 0.11-5.05 | N/A | N/A |

Note that N/A (Not Available), M(Male), and F(Female). In the "Algorithm" column, C(Classification), R(Regression), and H(Hybrid). In the "CS (percent)" column, the tolerable absolute error is set as 10 years.

- During the estimation process, the feature extraction is the first important step which extremely effects on a learning approach and its obtained results. The second influential step of an age estimation technique is training of pattern recognition approach based on the extracted features vectors. It is obvious that the influence of the feature extraction and training steps should be considered during the age estimation process.
- As depicted in Table.1, the highest performance rate is achieved by the HWCM due to using a combination of Haar wavelet transform and color moment methods to extract feature elements. This method employed a SVR model in order to improve the training model for facial aging. In addition, the second-high performance rate is obtained by SVM+, which utilizes a ranking SVM based on attributes during the feature extraction step. Also, it employs the ranking scores to group features and improve the accuracy of age estimation during the training process by controlling outliers in the training datasets.
- Using a combination of a SVM based feature extraction and a SVR model during the training process might improve more the accuracy and performance of age estimation techniques.
- As a result of the HWCM, using a combination of Haar wavelet transform and color moment methods during the feature extraction can be combined by the SVM during the training process for achieving better performance.
- A promising technique to age estimation is to combine regression, classification, and attribute learning techniques as provided in [27], [46], [47]. advanced schemes for the combination of classifiers and regressors, ranking attribute learning so that the accuracy of age estimation might be improved further.

## 4. Conclusion

This study presents an overview of the state-of-the-art techniques for age synthesis and estimation via facial images, which have become popular in recent years due to their promising real-world applications in many emerging fields. During the past two decades, the explosively comprehensive efforts from both industry and academic researchers have been done for modeling facial images and designing age estimation techniques, collecting facial aging datasets, and improving the performance evaluation with valid protocols. Moreover, different solutions to technical challenges have been presented by computer vision researchers. In this case study, first of all, we have overviewed the existing facial age synthesis and age estimation techniques. Secondly, we have outlined some highlight techniques and their performance rates. Finally, we have suggested some of the guidelines and directions that could merit further attention in future works.